\documentclass[runningheads]{llncs}
\usepackage[T1]{fontenc}
\usepackage{url}
\usepackage{subcaption}
\let\llncssubparagraph\subparagraph
\let\subparagraph\paragraph
\usepackage[compact]{titlesec}
\let\subparagraph\llncssubparagraph

\usepackage{graphicx}
\begin{document}
\title{Immersive 3D Simulator for Drone-as-a-Service}
%
%
\author{
Jiamin Lin \and
Balsam Alkouz \and
Athman Bouguettaya \and
Amani Abusafia
}

\institute{University of Sydney, Australia \\
\email{jlin6645@uni.sydney.edu.au},
\email{\{balsam.alkouz,athman.bouguettaya, amani.abusafia\}@sydney.edu.au}}
\maketitle              
\begin{abstract}
We propose a 3D simulator tailored for the Drone-as-a-Service framework. The simulator enables employing dynamic algorithms for addressing realistic delivery scenarios. We present the simulator's architectural design and its use of an energy consumption model for drone deliveries. We introduce two primary operational modes within the simulator: the edit mode and the runtime mode. Beyond its simulation capabilities, our simulator serves as a valuable data collection resource, facilitating the creation of datasets through simulated scenarios. Our simulator empowers researchers by providing an intuitive platform to visualize and interact with delivery environments. Moreover, it enables rigorous algorithm testing in a safe simulation setting, thus obviating the need for real-world drone deployments. Demo: \url{https://youtu.be/HOLfo1JiFJ0}


\keywords{Unmanned Aerial Vehicles \and Drones \and Drone Delivery \and Simulator.}
\end{abstract}
\section{Introduction}

The rapid proliferation of Unmanned Aerial Vehicles (UAVs), also known as drones, has transformed many industries, including logistics and transportation \cite{eskandaripour2023last}. The potential for drone delivery systems to enhance the efficiency and speed of last-mile deliveries has captured the attention of both researchers and industry leaders \cite{eskandaripour2023last}\cite{alkouz2020formation}. In parallel, the concept of \textit{skyway networks}, i.e., aerial corridors designated for autonomous drone traffic, has gained traction as a solution to the challenges of managing and regulating drone operations in urban environments \cite{shahzaad2023optimizing}. Consequently, several drone service approaches have been proposed to achieve efficient delivery \cite{shahzaad2023optimizing}\cite{alkouz2020swarm}. However, testing these approaches by deploying them on physical drones is challenging due to the potential safety risks \cite{shahzaad2023optimizing}.  
Thus, there is a pressing need for \textit{advanced simulators} to harness the full potential of drone delivery within skyway networks. These tools should be capable of modeling, analyzing, and optimizing these complex systems, especially within the framework of \textit{Drone-as-a-Service (DaaS)}.\looseness=-1

Traditional simulations often rely on numerical tracking and data-driven modeling \cite{alkouz2021reinforcement}\cite{janszen2021constraint}\cite{bradley2023service}. However, these may fall short of replicating the dynamics of real-world urban drone operations. 3D simulation, on the other hand, provides a holistic and \textit{immersive} environment that replicates the physical aspects of the skyway network. This offers a more accurate representation of how drones interact with their surroundings. Moreover, it allows for the evaluation of spatial relationships, environmental factors, and dynamic obstacles, all of which are critical in urban airspace management \cite{alkouz2021service}. In a 3D simulation, researchers and stakeholders may visualize complex scenarios, observe emergent behaviors, and validate the feasibility of their solutions. 

The existing drone simulation tools often focus on specific aspects of drone operations, e.g., obstacle avoidance \cite{garcia2022simulation}.  However, they lack the integration required for comprehensive urban delivery system evaluation. Notable tools include AirSim \cite{shah2018airsim}, Gazebo \cite{garcia2022simulation}, and PX4 \cite{garcia2022simulation}. These tools focus on modeling drone flight dynamics and sensor simulations. However, they primarily serve the needs of the drone development community and \textit{are hardly useful in addressing the unique challenges of urban drone delivery within a skyway network}.

We propose a 3D simulator that distinguishes itself by offering a comprehensive analytical and visual framework. Our simulator is \textit{tailored to the DaaS framework within urban environments} \cite{lee2022autonomous}. It encompasses \textit{route planning, real-world data integration, airspace management, and scalability analysis}, all in a 3D environment that replicates the complexities of skyway networks. This comprehensive approach aligns with the growing demand for seamless, end-to-end solutions. It enables researchers, businesses, city planners, and drone operators to evaluate and optimize their operations from a service-oriented perspective.

\section{Demo Setup}
Drone-as-a-Service (DaaS) for delivery refers to the concept of providing delivery services within a skyway environment \cite{shahzaad2023optimizing}\cite{lee2021package}. The skyway network comprises building rooftops equipped with charging and landing pads, functioning as nodes in the system. Any segment of the network that is served by a delivery drone is categorized as a DaaS service. The concept of Swarm-based Drone-as-a-Service (SDaaS) extends this idea to encompass delivery services provided by a swarm of drones \cite{guo2023drone}\cite{alkouz2022density}. In this context, a segment in the network serviced by a drone swarm carrying multiple packages is referred to as an SDaaS segment \cite{liu2022constraint}\cite{alkouz2021provider}. Fig. \ref{fig:editMode} provides a visual representation of the skyway arrangement.

This demo paper showcases a simulation system specifically designed for evaluating DaaS and SDaaS within a 3D simulated skyway environment. In this section, we will delve into the key components that constitute this simulator.

\subsection{System Architecture}

\begin{figure}[htbp!]
    \centering
    \includegraphics[width=\textwidth]{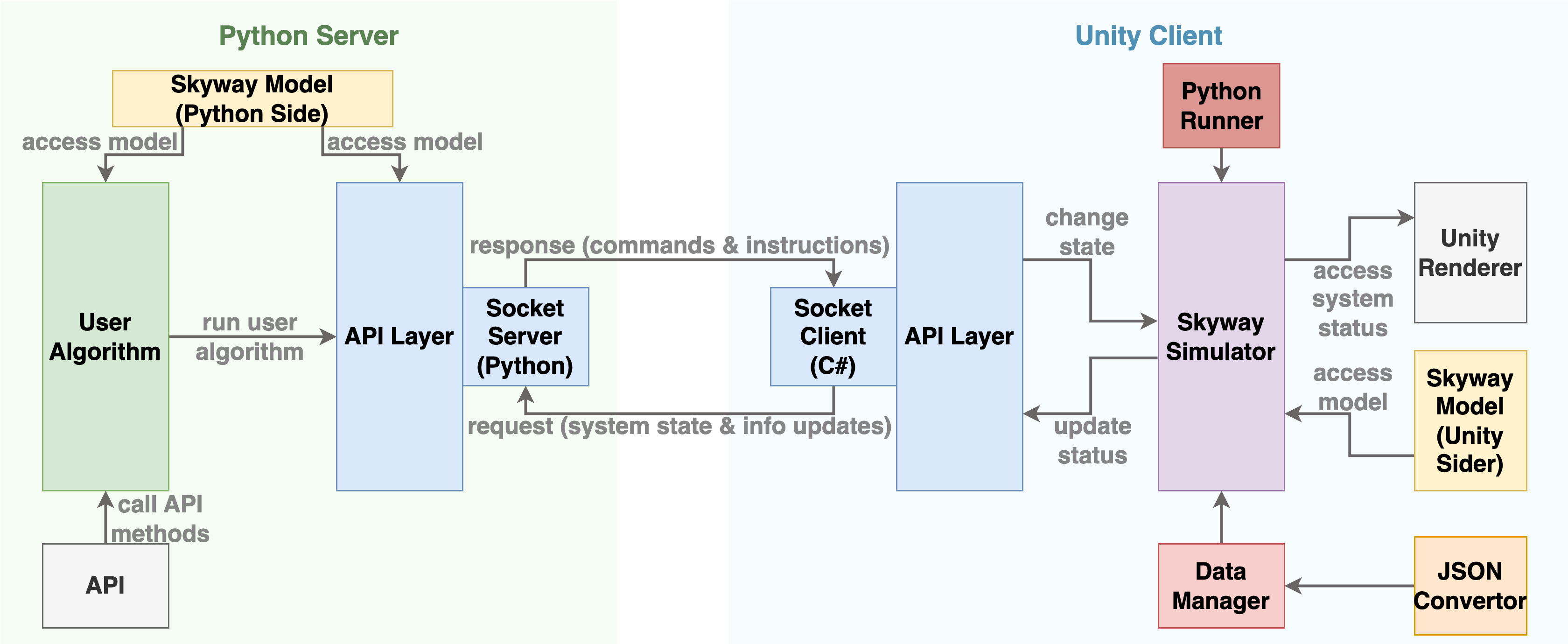}
    \caption{System Architecture}
    \label{fig:architecture}
\end{figure}

The system architecture comprises two primary modules that collaborate closely. These modules consist of a front end (client side) and a back end (server side) (See Fig. \ref{fig:architecture}). The front end, implemented in Unity\footnote{\url{https://unity.com/}}, serves as the visual and simulation hub where users interact with the system. Meanwhile, the back end is a Python script environment, chosen due to its popularity among researchers \cite{shahzaad2023optimizing}\cite{alkouz2023failure}. Users may incorporate their Python code to implement their algorithms in the back end and visualize the simulation at the front end.

Within the Unity-based front end, users engage with the Skyway Simulator and Skyway Model, allowing them to interact with the system's simulation. Changes made in the front end are transmitted to the back end as drones traverse the network. At each node, the Unity client updates the Python server, providing the algorithm with real-time information about the Skyway's status. The algorithm reacts dynamically to any changes, composing services as necessary. It is essential to emphasize that this architecture operates non-deterministically, and updates are regularly exchanged to trigger actions between the client and server sides.\looseness=-1

\subsection{Energy Consumption Model}
We employ the Kirchstein energy consumption model to simulate the drones energy usage\cite{kirschstein2020comparison}. This model offers a distinct advantage as it accounts for drone flight's ascent and descent phases. Given that our simulation involves connecting various buildings of varying heights with skyway segments, it becomes crucial to factor in the flight angle as drones navigate this terrain. The model accounts for vertical movements and hovering This is essential for accurately depicting the drones' actions during takeoff and landing at designated recharging pads. Furthermore, it is important to highlight that this model is theoretical, which makes it more suitable for simulations and adaptable to various drone types. This sets it apart from regression models that depend on empirical data from actual drones \cite{dorling2016vehicle}. 


\subsection{Simulation Environment}
For the simulation environment, we incorporated a 3D city model in Unity to represent the buildings. Our simulator offers two primary modes of operation. The first is the \textit{Edit Mode}, which allows users to customize the skyway environment and edit the experimental variables according to their preferences. The second mode is the \textit{Runtime Mode}, during which drones initiate flights based on the Python algorithm provided, facilitating data exchange between the client and server sides at each node. Below, we provide an overview of the specifics for each mode:
\subsubsection{Edit Mode:}


\begin{figure}
    \centering
    \includegraphics[width=\linewidth]{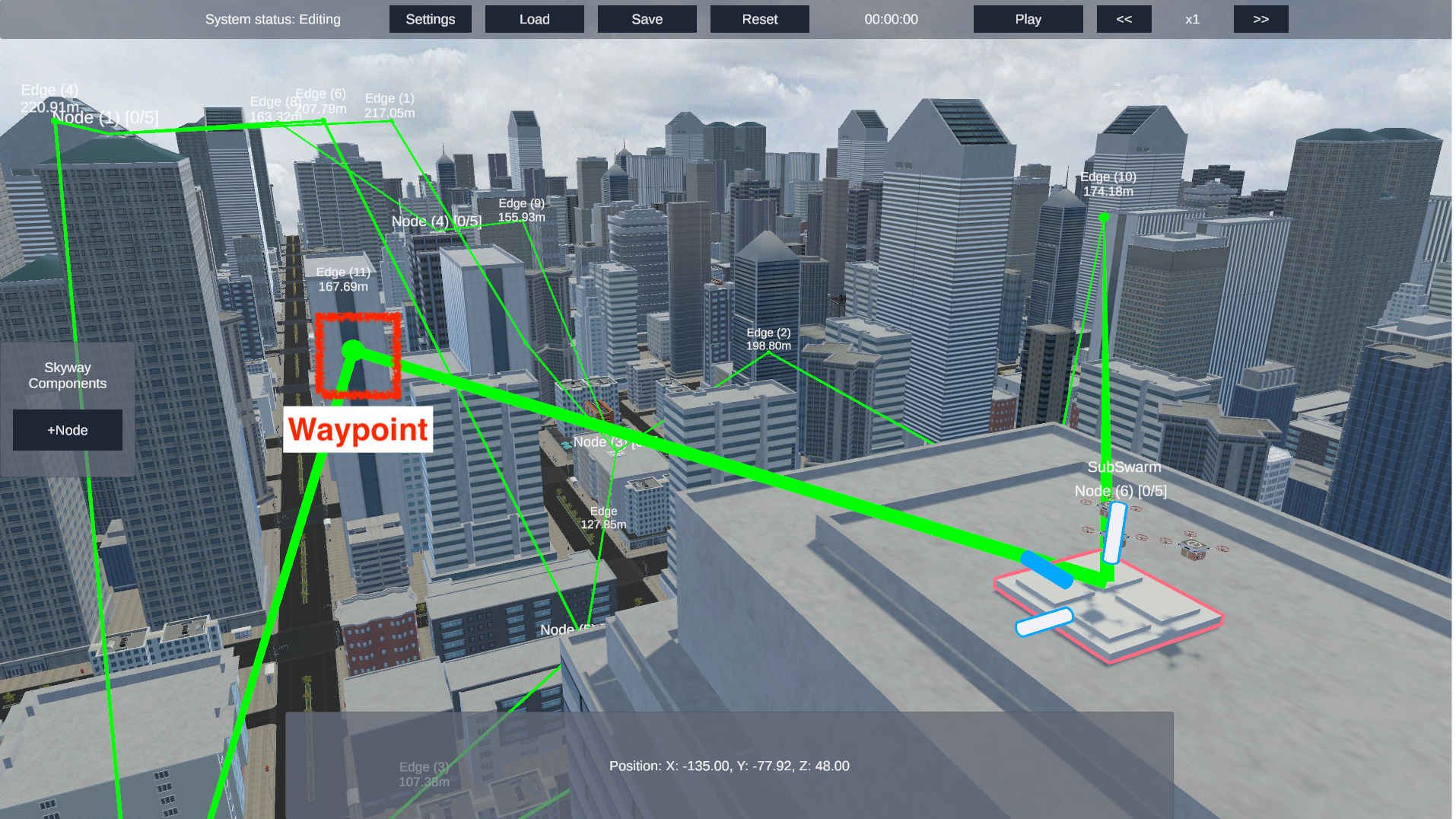}
    \caption{Edit Mode}
    \label{fig:editMode}
\end{figure}

\begin{figure}
    \centering
    \includegraphics[width=\linewidth]{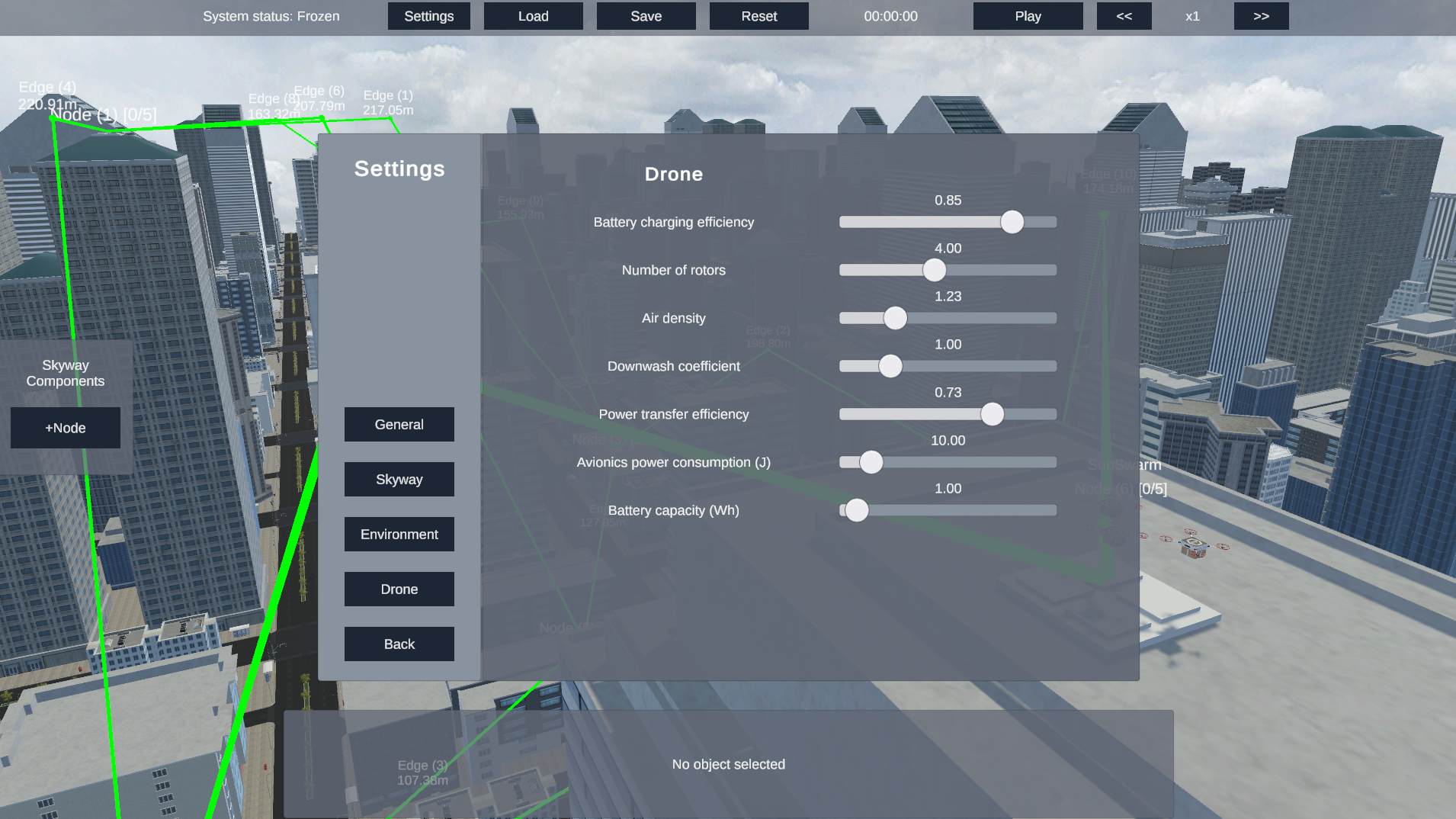}
    \caption{Settings in Edit Mode}
    \label{fig:settings}
\end{figure}
Illustrated in Fig. \ref{fig:editMode}, the edit mode provides users with the ability to perform various actions such as adding or removing nodes, connecting or disconnecting segments, and resizing the skyway network to suit the requirements of their experiments. This functionality proves invaluable for modeling both densely interconnected and sparsely connected networks. Additionally, users can reposition nodes through intuitive mouse controls within this mode. The "Load" button facilitates the importation of a JSON file containing essential network setup details. This includes node configurations, their positions, and segment specifications. Conversely, the "Save" button empowers users to preserve the current network configuration as a JSON file for future reference and utilization.\looseness=-1 

Furthermore, we introduce the concept of \textit{waypoints}, as depicted in Fig. \ref{fig:editMode}. Waypoints connect segments that cannot be directly linked to two nodes due to disparities in elevation or challenging terrain. 
This concept adds a layer of realism by acknowledging that drones cannot always establish a direct, unobstructed flight path between nodes \cite{alkouz2022flight}. Lastly, when the settings button is pressed, users can specify attributes related to the energy consumption model, payload capacity, and drone speed (Fig. \ref{fig:settings}).



\subsubsection{Runtime Mode:}
\begin{figure}[htbp!]
    \centering
    \includegraphics[width=\textwidth]{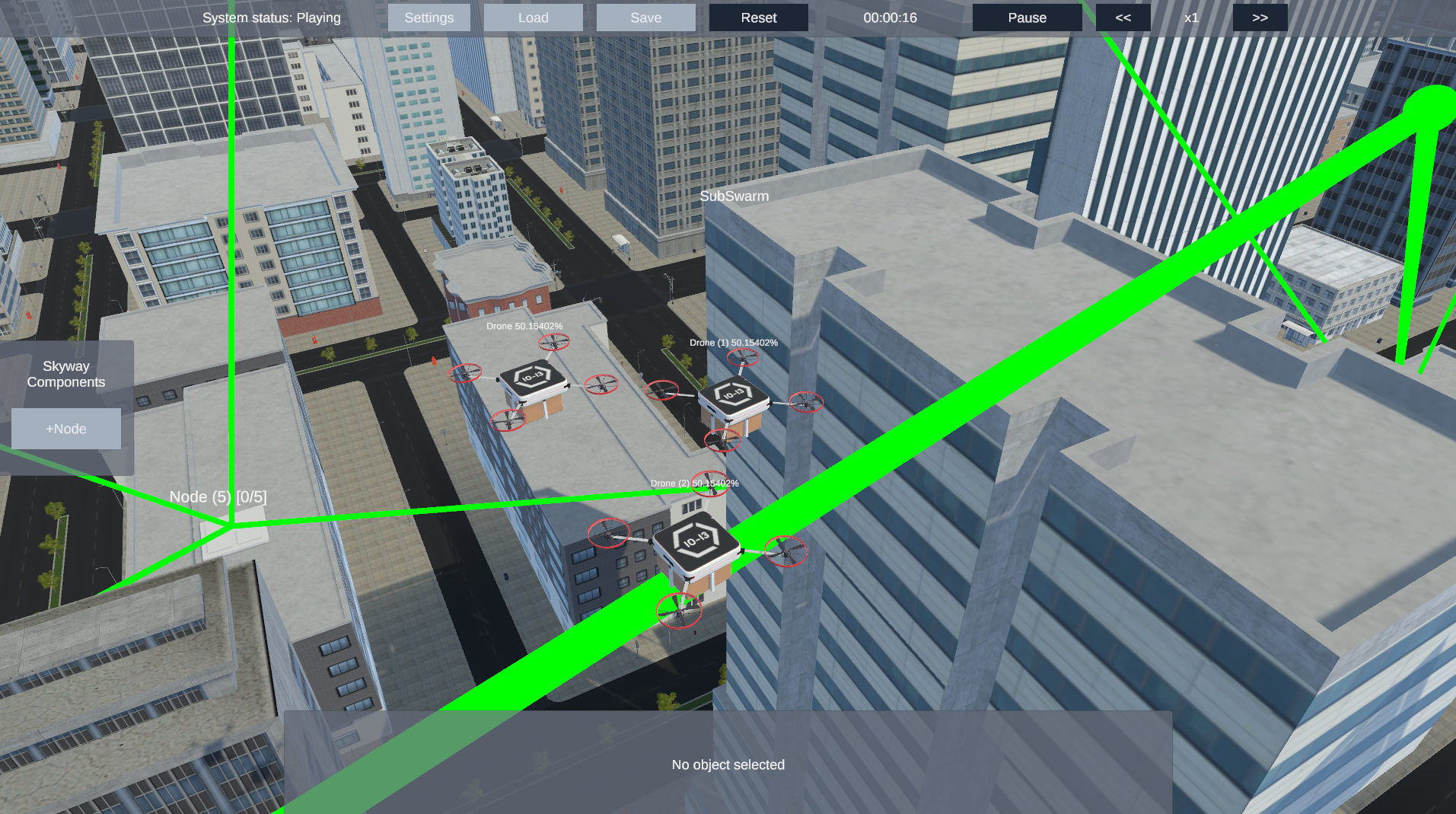}
    \caption{Runtime Mode}
    \label{fig:runtime}
\end{figure}

Once the play button is hit in the Edit mode, the Runtime mode starts (Fig. \ref{fig:runtime}). In the runtime mode, the drones become visible and begin traversing according to the provided algorithm. As previously mentioned, this tool operates in a non-deterministic manner. Consequently, if a segment (a service) becomes inaccessible from the originally composed path, the algorithm promptly adapts to this change, redirecting the drones accordingly.\looseness=-1


The camera tracks the drones during their journey automatically. However, users also have the option to take control using the WASD buttons for manual camera adjustments. To provide users with an enriched experience, double-clicking on any item in the environment offers a close-up view. Each object within the simulation is thoughtfully labeled to convey its relevant attributes, such as segment length or drone battery percentage. The timer displays the simulation time elapsed since the commencement of the trip and permits users to adjust the simulation speed, enabling the drones to move at varying rates.


\subsection{Data Collection}

This tool serves a dual purpose: not only it assists users in visualizing potential scenarios in drone deliveries under the DaaS model but also functions as a valuable data collection tool for environmental interactions. Following each trip, users may export a CSV file containing comprehensive data regarding the status of every drone in the network. This data encompasses crucial information such as drone energy consumption, battery levels, time spent at each node, and travel times for each segment at each time frame. Such data can be leveraged to refine and improve DaaS composition algorithms.




\subsubsection{Acknowledgements} This research was partly made possible by LE220100078 and DP220101823 grants from the Australian Research Council. The statements made herein are solely the responsibility of the authors.

\bibliographystyle{splncs04}
\bibliography{icsoc}

\newpage
\section*{Demonstration Requirements}
This demonstration will showcase the 3D simulator and provide attendees with the opportunity to test and interact with it. We will utilize our laptops for this purpose. To facilitate the demo, a standard booth set up at the conference venue is necessary with electricity outlets. Having a screen projector is preferable to enhance visibility for the audience.

\end{document}